\documentclass[journal]{IEEEtran}
\IEEEoverridecommandlockouts
\usepackage{cite}
\usepackage{amsmath,amssymb,amsfonts}
\usepackage{algorithmic}
\usepackage{graphicx}
\usepackage{textcomp}
\usepackage{xcolor}

\usepackage{tabularx}

\usepackage{mdwlist}
\usepackage[ruled,vlined,shortend, linesnumbered]{algorithm2e} 
\usepackage[]{algorithmic} 
\usepackage{booktabs}
\usepackage{color}

\usepackage{amsthm,amssymb}
\usepackage{amsmath}
\usepackage{amsfonts}
\usepackage{bm} 
\usepackage{multirow}
\usepackage{mathrsfs}
\usepackage{amsbsy}
\usepackage[mathscr]{eucal} 
\usepackage[flushleft]{threeparttable}
\usepackage[hidelinks]{hyperref}
\usepackage{comment}
\usepackage{paralist}
\usepackage{pifont}

\usepackage{bbding} 

\def\BibTeX{{\rm B\kern-.05em{\sc i\kern-.025em b}\kern-.08em
    T\kern-.1667em\lower.7ex\hbox{E}\kern-.125emX}}

\newcommand{\hide}[1]{}

\newcommand{\bit}{\begin{compactitem}}
\newcommand{\eit}{\end{compactitem}}
\newcommand{\ben}{\begin{compactenum}}
\newcommand{\een}{\end{compactenum}}

\usepackage{xcolor}

\usepackage[colorinlistoftodos]{todonotes}
\usepackage{xcolor}

\theoremstyle{definition}

\usepackage{amsthm}
\usepackage{amsmath}
\usepackage{amssymb}
\usepackage{thmtools}
\usepackage{thm-restate}
\usepackage{hyperref}

\usepackage{mathtools}

\begin{document}

\title{Towards Practical Physics-Informed ML Design and Evaluation for  Power Grid
}

\author{Shimiao~Li,~\IEEEmembership{Graduate Student Member,~IEEE,}
	Amritanshu~Pandey,~\IEEEmembership{Member,~IEEE,}
	and~Larry~Pileggi,~\IEEEmembership{Fellow,~IEEE}
\thanks{}
}

\maketitle

\begin{abstract}
	When applied to a real-world safety critical system like the power grid, general machine learning methods suffer from expensive training, non-physical solutions, and limited interpretability. 
To address these challenges for power grids, many recent works have explored the inclusion of grid \textit{physics} (i.e., domain expertise) into their method design, primarily through including system constraints and technical limits, reducing search space and defining  meaningful features in latent space. Yet, there is no general methodology to evaluate the practicality of these approaches in power grid tasks, and limitations exist regarding scalability, generalization, interpretability, etc. This work formalizes a new concept of \textit{physical interpretability} which assesses  
\textit{how a ML model makes predictions in a physically meaningful way}
and introduces an evaluation methodology that identifies a set of attributes that a practical method should satisfy. 
Inspired by the evaluation attributes, the paper further develops a novel contingency analysis warm starter for MadIoT cyberattack, based on a conditional Gaussian random field. 
This method serves as an instance of an ML model that can incorporate diverse domain knowledge and improve on these identified attributes. Experiments validate that the warm starter significantly boosts the efficiency of contingency analysis for MadIoT attack even with shallow NN architectures.

\end{abstract}


\section{Introduction}
\label{sec:Introduction}
Advances in data collection and storage have increased the grid data exponentially. Consequently, a vast literature of data-driven approaches to aid or replace the existing data analytics for power system operation, control and planning has emerged in the context of growing uncertainties and threats. However, when used on a real physical system like the power grid, a blind application of machine learning suffers from critical limitations. On the one hand, massive training data are required to define the complicated feature space and the possible scenarios. On the other hand, as a black box, general machine learning tools raise the fear (and the actual problem) of generating infeasible or meaningless outputs on unseen data due to limited interpretability.

To address critical challenges with general ML methods, recent advancements have been made to include \textit{grid physics}, also called domain knowledge or expertise, into the general ML methods. These physics-informed ML methods can be broadly categorized as follows (a detailed discussion is in Section \ref{sec:ML survey}):
\begin{itemize}
    \item reducing the search space \cite{rl-topology}\cite{rl-multiagent-v}\cite{rl-multiagent-shunt}
    \item enforcing system constraints and technical limits \cite{ppf-dnn}\cite{pf-dnn-topo}\cite{gnn-pf}\cite{dcopf-dnn}\cite{acopf-dnn-donti}\cite{gnn-acopf}\cite{gnn-acopf-warm}\cite{unrolled-se-2018}\cite{unrolled-se}\cite{unrolled-gnu-se}
    \item crafting meaningful features and defining an interpretable latent  space \cite{pmu-graph-temporal}\cite{pmu-graph-spatial}\cite{cascade-influence-model}\cite{dynwatch}
\end{itemize}

While these physics-informed ML methods address challenges with general ML methods, critical limitations still exist regarding scalability\cite{scalability}, generalization\cite{generalization-DL}, interpretability\cite{interpretability}, etc., especially in methods designed with insufficient knowledge of the physical system. Firstly, many of these methods do not consider the dynamic nature of grid conditions, in terms of network topology, loads, and generations, requiring their model to be retrained upon any changes. Secondly, they do not fully address the needs of safety critical power grid tasks. For instance, reducing false negative and variance (which
may often be neglected) is more critical than reducing bias. 
Moreover, regarding physical interpretability\cite{interpretability}, some methods have limited physical meaningfulness. For instance, it is difficult to interpret, from a physical perspective, why a model like random forest achieves performance comparable to, or even better than, DC power flow. Therefore, while many methods perform well on training datasets, there is minimal evidence that these methods can generalize. 
We will discuss more on the limitations in Section \ref{sec:ML survey}.

These limitations largely come from a lack of power grid-based specifications for ML models' design and evaluation. As such, this paper introduces an evaluation methodology that summarizes, from power system perspective, some important attributes that are necessary for ML methods to be practical on grid-specific tasks.  Among the identified attributes, this paper defines a new concept of \textit{physical interpretability} which considers a method's physical meaningfulness, which is missing from the general field of ML.

Further, guided and inspired by these attributes, this paper proposes a novel contingency analysis\cite{contingency-analysis} warm starter based on pairwise conditional Gaussian Random Field (cGRF), in the context of MadIoT attack (a cyber attack on IoT-controlled loads, see Section \ref{sec:madiot}). The warm starter predicts the post-contingency voltages which are used as the initial condition in power flow analysis. The technique reduces the number of iterations needed to simulate contingency events due to the MadIoT cyberattack by averagely more than 70\% on a 2000-bus case.
In the development of the method, we integrate diverse \textit{grid physics} to improve on the attributes we identified in Section \ref{sec:framework}. Further in experiment results, we show that by integrating \textit{grid physics}, even a shallow neural network (NN) architecture can enable the proposed warm starter to significantly reduce the run-time of power simulations for large networks when evaluating contingencies corresponding to MadIoT cyberattacks.

\section{Related Work} \label{sec:relatedwork}
\subsection{Physics-informed ML for power grids: literature review}\label{sec:ML survey}

Many prior works have included domain-knowledge in their methods to address the problem of missing \textit{physics} in generic ML tools. These methods collectively fall under physics-informed ML paradigm for power grid operation, control, and planning and can be broadly categorized into following categories:

\subsubsection{Reducing search space}
In these methods domain knowledge is used to narrow down the search space of parameters and/or solutions. For example, \cite{rl-topology} designed a grid topology controller which combines reinforcement learning (RL) Q-values with power grid simulation to perform a 
\textit{physics}-guided action exploration, as an alternative to the traditional epsilon-greedy search strategy. Works in \cite{rl-multiagent-v}-\cite{rl-multiagent-shunt} studied the multi-agent RL-based power grid control. In these approaches the power grid is partitioned into  controllable sub-regions based on domain knowledge (i.e. electrical characteristics) to reduce the high-dimensional continuous action space into lower-dimension sub-spaces.

\subsubsection{Enforcing system constraints and technical limits}

Many recent works apply deep learning (mostly deep neural nets) for power grid analysis. These include but are not limited to power flow (PF)\cite{ppf-dnn}\cite{pf-dnn-topo}\cite{gnn-pf}, DC optimal power flow (DCOPF)\cite{dcopf-dnn}, ACOPF\cite{acopf-dnn-donti}\cite{gnn-acopf}\cite{gnn-acopf-warm} and state estimation (SE) \cite{unrolled-se-2018}\cite{unrolled-se}\cite{unrolled-gnu-se}. These methods are generally based on (supervised) learning of an input-to-solution mapping using historical system operational data or synthetic data. One type among these works are unrolled neural networks \cite{unrolled-se-2018}\cite{unrolled-se}\cite{unrolled-gnu-se} whose layers mimic the iterative updates to solve SE problems using first-order optimization methods, based on quadratic approximations of the original problem. 
There are also methods that learn the 'one-step' mapping function. 
Among these, some use deep neural network (DNN) architectures\cite{ppf-dnn}\cite{pf-dnn-topo}\cite{acopf-dnn-donti}\cite{dcopf-dnn} to learn high-dimensional input-output mappings, some use recurrent neural nets (RNNs)\cite{rnn-dse} to capture some dynamics, and some apply graph neural networks (GNN)\cite{gnn-pf}\cite{gnn-acopf}\cite{gnn-se} to capture the exact topological structure of power grid.
To promote \textit{physical} feasibility of the solution, many works impose equality or inequality system constraints by i) encoding hard constraints inside layers (e.g. sigmoid layer to encode technical limits of upper and lower bounds), ii) applying prior on the NN architecture (e.g., Hamiltonian\cite{hamiltonianNN} and Lagrangian
neural networks\cite{lagrangianNN}), iii) augmenting the objective function with penalty terms (supervised\cite{ppf-dnn} or unsupervised\cite{acopf-dnn-donti}\cite{pf-dnn-topo}), iv) projecting outputs \cite{dcopf-dnn} to the feasible domain, or v) combining many different strategies. 
In all these methods, incorporation of the (nonlinear) constraints remains a challenge, even with state-of-the-art toolboxes\cite{neuromancer2022}, and most popular strategies lack rigorous guarantees of nonlinear constraint satisfaction. 

While these methods have advanced the state-of-the-art in physics-informed ML for power grid applications, critical limitations in terms of generalization, interpretation, and scalability exist in existing methods. We further discuss them:

\textbf{Limited generalization:} Many existing methods no longer work well when certain grid attributes, like network topology, change. By design, graph model-based methods (e.g., GNN\cite{gnn-pf}\cite{gnn-acopf}\cite{gnn-se}) naturally embed the system connectivity to account for topology differences. Unfortunately, most non-graphical models, once trained, only work for one fixed topology and cannot generalize to dynamic grid conditions. Work in \cite{pf-dnn-topo} encoded topology information into the penalty term through the use of admittance and adjacency matrix. The method turns out to have better topology adaptiveness, yet the use of the penalty term fails to rigorously encode the exact topology. Similarly, work in \cite{unrolled-gnu-se} accounted for topology in NN, inexactly and implicitly, by applying a topology-based prior, which is implemented by a penalty term.  

\textbf{Limited interpretability:} Despite that many machine learning tools (NN, decision trees, K-nearest-neighbors) are universal approximators, interpretations of their functionality from a physically meaningful perspective are still very limited. Unrolled neural networks whose layers mimic the physical solvers are more decomposable and interpretable, yet so far in literature, they are only applicable to unconstrained problems like SE. GNN based models\cite{gnn-pf}\cite{gnn-acopf}\cite{gnn-se} which naturally represent the grid topology enables better interpretability in terms of graph structure, yet their final predictions, which are typically obtained from hidden features and node embedding within a local neighborhood, lacks global considerations. \cite{ppf-dnn} provides some interpretation of its DNN model for PF, by matching the gradients with power sensitivities and this finding enables accelerating the training by pruning out unimportant gradients. \cite{pf-dnn-topo} learns a weight matrix that can approximate the bus admittance matrix; however, with only limited precision. Nonetheless, at present there is no standard measure to assess the degree of physical meaningfulness for various physics-informed ML methods. Compared with the purely physics-based models (physical equations) and constraints, most ML models still have a lot of opacity and blackbox-ness, raising the fear of outputting bad predictions on unseen data. 

\textbf{Scalability issues}:
In the case of large-scale systems, models that learn the mapping from high-dimensional input-output pairs will inevitably require larger and deeper designs of model architecture, and thereafter massive data to learn such mappings. This is a serious drawback for practical use in real-world power grid analytics.

\subsubsection{Extracting meaningful features or crafting an interpretable latent space} 
Many works exist in this class. For example, \cite{pmu-graph-temporal}\cite{pmu-graph-spatial} learned the latent representation of sensor data in a graph to capture temporal dependency  \cite{pmu-graph-temporal} or spatial sensor interactions \cite{pmu-graph-spatial}.
\cite{cascade-influence-model} applied influence model to learn, for all edge pairs, the pairwise influence matrices which are then used for the prediction of line cascading outages. \cite{dynwatch} crafted a graph similarity measure from power sensitivity factors, and enables anomaly detection in the context of changing topology by weighing historical data based on these similarities.

\subsection{MadIoT: IoT-based Power Grid Cyberattack}\label{sec:madiot}

\begin{figure}[h]
	\centering
	\includegraphics[width=1.0
	\linewidth]{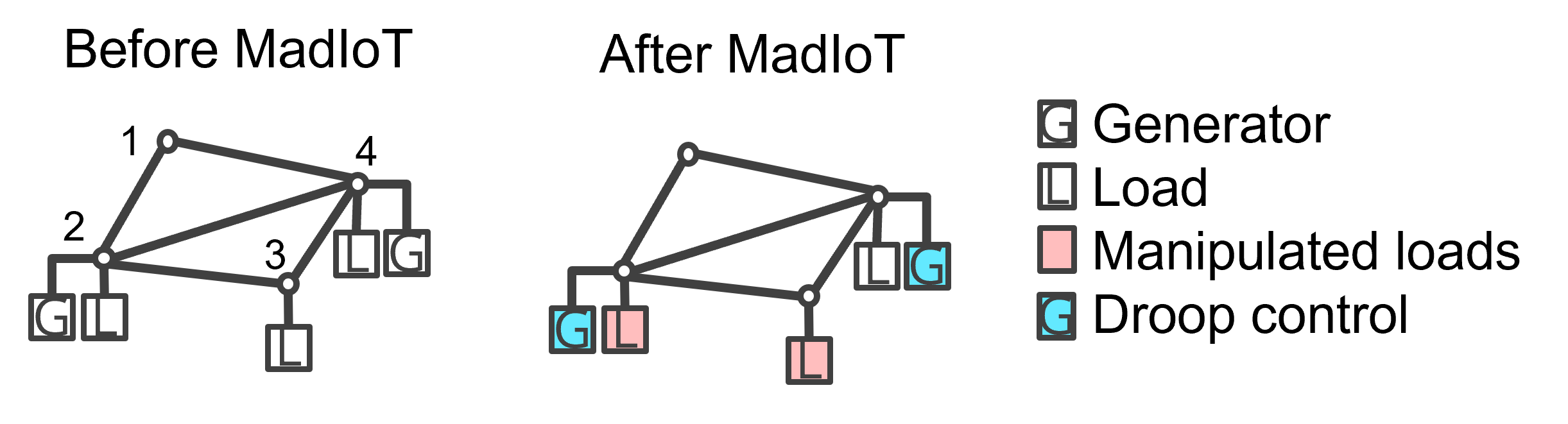}
	\caption[]{A toy example of MadIoT: a subset of loads are manipulated; to consider steady state impact, we simulate droop control which adjusts generators to compensate for some demand change.}
	\label{fig:madiot}
\end{figure}

The proliferation of IoT devices has raised concerns about IoT enabled cyber attacks. \cite{madiot} proposed a threat model, namely \textit{BlackIoT} or \textit{MadIoT} attack, where an attacker can manipulate the power demand by synchronously turning on/off, or scaling up/down some high-wattage
IoT-controlled loads on the grid. This can lead to grid instability or grid collapse. Figure \ref{fig:madiot} illustrates an attack instance of the MadIoT threat. To evaluate the impact of a postulated attack, \cite{madiot} uses  steady-state analysis of the power grid while taking into account the droop control, protective relaying, and thermal and voltage limits for various components.

\subsection{Contingency analysis}\label{sec:pf and ca}
Contingency Analysis (CA)\cite{contingency-analysis} is a "what if" simulation used in both operation and planning to evaluate the line flow and voltage impacts of possible disturbances and outages. Today during operation, real-time CA is performed approximately every 5-30 minutes (5min in ERCOT\cite{ercot-rtca-5min}, at least 30min required by NERC). These CA runs simulate a predefined set of N-1 (loss of one element) contingencies. In near future, with potential likelihood of cyberattacks (e.g., MadIoT), the power grid will have to be evaluated under a new set of cyberattack-based contingencies for reliable and secure operation. Instead of  traditional N-1 contingencies, these contingencies will represent N-x events (simultaneous loss of x components).
While fast evaluation of many N-1 contingencies is possible due to close proximity of final solutions to initial conditions, these characteristics no longer hold true for cyberattack-based N-x contingencies. Therefore, it may be a challenge to evaluate them in a limited time.

To address this challenge, one option is to use effective contingency screening \cite{sugar-lodf} strategies that can significantly reduce the number of contingencies to simulate. Another option is to have a warm starter (i.e., improved initial conditions) that can speed up the convergence of each simulation. This paper proposes a method for the latter.

\section{Proposed Evaluation Framework}

In this section, we first define a new concept of \textit{physical interpretability} to measure a model's physical meaningfulness and then identify a set of attributes that can qualitatively and quantitatively measure the goodness of a physics-informed ML method for power grid application. From the authors' viewpoint, methods' practicality (for power grids) can be comprehensively described via the identified attributes in the following section.

\subsection{Physical Interpretability}\label{sec:interpretability}

\setlength{\tabcolsep}{6pt}
\begin{table*}[h]
\small
\centering
\caption{Dimensions of physical Interpretability}
\label{tab:physical interpretability}
 \begin{tabular}{llll}
 \toprule
 & & \multicolumn{2}{c}{Precision of physics integration}\\
  &\multicolumn{3}{c}{
 \begin{tabular}{ m{1em}m{7.5cm} m{7.5cm}}
   \rotatebox[origin=c]{90}{}
   & \setlength\parindent{64pt}
Exact representation
   & \setlength\parindent{64pt} Approximate representation  \\
\end{tabular}
 }\\
 \midrule
 \parbox[t]{2mm}{\multirow{2}{*}{\rotatebox[origin=c]{90}{Scale of physics integration}}} 
 &\multicolumn{3}{c}{
 \begin{tabular}{ m{1em}|m{7.5cm}|m{7.5cm} }
  \rotatebox[origin=c]{90}{Global}
   & \begin{itemize}
    \item Mapping output onto feasible region to enforce rigorous satisfaction of complete system constraints:
    \begin{itemize}
        \item (during training) gradient-based reprojection to feasible region \cite{acopf-dnn-donti}
        \item (post processing) passing outputs to a physical equation solver  \cite{dcopf-dnn}
    \end{itemize}
    \item Graphical model (e.g. graph neural nets) to capture exact system topology\cite{gnn-pf}\cite{gnn-acopf-warm}\cite{gnn-acopf}\cite{gnn-se}
   \end{itemize}
   & \begin{itemize}
    \item Unrolling the iterative physical solvers inside deep learning layers \cite{unrolled-se-2018} \cite{unrolled-se}\cite{unrolled-gnu-se}
       \item Approximating complete system constraints through  penalty terms in objective function
       \begin{itemize}
           \item supervised penalty\cite{ppf-dnn}
           \item unsupervised penalty \cite{acopf-dnn-donti}\cite{opf-rl}\cite{pf-dnn-topo}
       \end{itemize}
       \item System-level proxy given by ML architecture or parameters: 
       \begin{itemize}
        \item Linear system-level approximation\cite{pf-dnn-topo}
        \item Nonlinear system-level approximation \cite{ppf-dnn}
    \end{itemize}
   \end{itemize}  \\
  \cline{2-3}
\end{tabular}
 }\\
 &\multicolumn{3}{c}{
 \begin{tabular}{ m{1em}|m{7.5cm}|m{7.5cm} }
  \rotatebox[origin=c]{90}{Partial    }
   & \begin{itemize}
    \item Implementation of partial/local physical constraints and technical limits in ML architecture
    \item Enforcing partial constraints and limits by post-processing
   \end{itemize}
   & \begin{itemize}
       \item Approximating partial system constraints by penalty or prior
    \item Crafting interpretable hidden features and latent space (that encode physical characteristics) \cite{pmu-graph-temporal}\cite{pmu-graph-spatial}\cite{cascade-influence-model}\cite{dynwatch}
   \end{itemize}  \\
\end{tabular}
 }\\
 \bottomrule
 \end{tabular}
 \end{table*}
 
The field of general \textit{model interpretability}\cite{interpretability} mainly focuses on \textit{how does a model work}, to mitigate fears of the unknown by building trust, transferability, causality, etc.
Broadly, investigations of interpretability can be categorized into transparency (also called ad-hoc interpretations) and post-hoc interpretations. The former aims to elucidate the mechanism by which the ML \textit{blackbox} works before any training begins, by considering the notions of \textit{simulatability} (Can a human work through the model from input to output, in reasonable time steps through every calculation required to
produce a prediction?), \textit{decomposability} (Can we attach intuitive explanation to each parts of a model: each input, parameter, and calculation?), and \textit{algorithmic transparency} (Does the learning algorithm itself confer guarantees on convergence, error surface even for unseen data/problems?). Whereas \textit{post-hoc interpretation} aims to inspect a learned model after training, by considering its natural language explanations, visualizations
of learned representations/models, or explanations by
example.  A detailed overview can be found in \cite{interpretability}.

However, none of these concepts formally considers the physical meaningfulness of a ML model when it is applied on a physical system like the power grid. This raises a concern about limited interpretability as we discussed in Section \ref{sec:ML survey}.

To address this issue, this paper provides an extension of existing definitions in Section \ref{sec:ML survey} by formalizing a new concept of \textbf{\textit{physical interpretability - how a model makes predictions in a physically meaningful way.}} This new concept can measure \textbf{different levels} of physical interpretability by two attributes: \textit{scale} and \textit{precision}:
\begin{itemize}
    \item \textbf{scale} considers \textit{at what scale} is the domain knowledge applied to the ML model. The domain knowledge can be either a \textit{global} system physics or \textit{partial} system physics. For instance, with access to only partial system information or data (e.g. lack of observability), one can only integrate some physics to describe the partial or local system characteristics.
    \item \textbf{precision} considers \textit{to what precision} does the method include the  physics: it can be either an \textit{exact} representation of physics, or an \textit{approximate} representation of physics. 
\end{itemize}

Table \ref{tab:physical interpretability} summarizes  various levels of physical interpretability for existing physics-informed ML methods. 

\subsection{An Evaluation Methodology\label{sec:framework}}
A method's \textit{practicality} - \textit{is it a practical method for real-sized systems and real-world challenges?} is extremely pertinent when applying a ML model on a power grid, yet it is a multi-dimensional abstract concept that is hard-to-measure. 

\begin{figure}[h]
	\centering
	\includegraphics[width=0.9\linewidth]{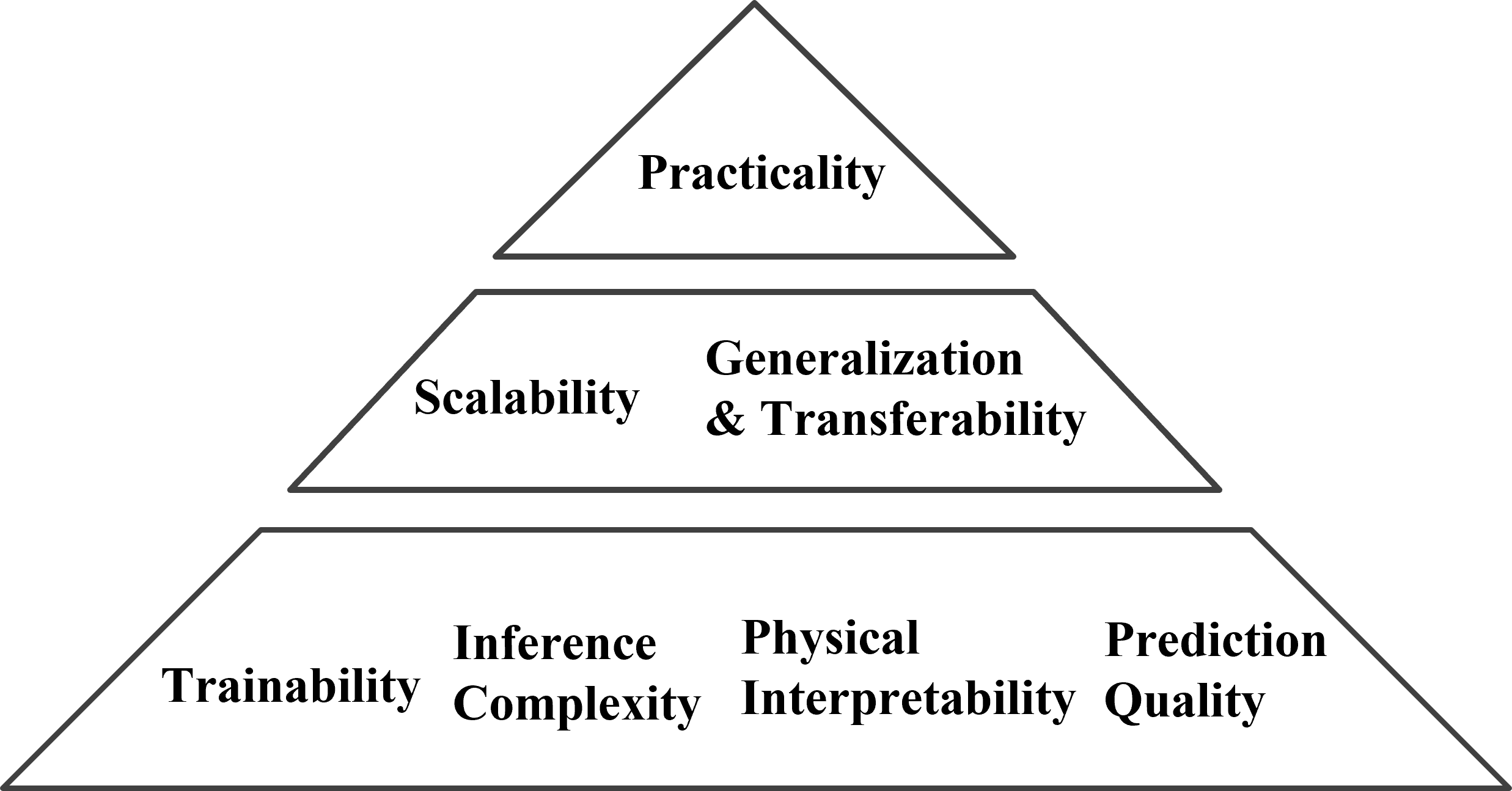}
	\caption[]{A evaluation methodology for physics-informed ML methods. It starts from basic attributes to higher-level abstract attributes}
	\label{fig: evaluation}
\end{figure}

This paper constructs an evaluation methodology, as shown in Figure \ref{fig: evaluation}, that summarizes some important attributes (ranging from basic attributes to higher-level abstract attributes) to evaluate the \textit{practicality} of a ML method. These attributes can qualitatively and quantitatively evaluate a ML model. To apply on grid-specific tasks, our evaluation methodology provides grid-specific explanations to each attribute. Within these attributes, we include a new concept of \textit{physical interpretability} that is proposed in Section \ref{sec:interpretability}. We describe the attributes in more detail following a bottom-up approach, beginning from basic attributes.

\subsubsection{Layer 1: Basic Attributes} The evaluation methodology starts from basic attributes that are easy to measure or quantify. 

\begin{itemize}
    \item \textit{trainability \cite{evaluation-metrics}- is it easy to train on power system data?} This is quantifiable by model complexity, i.e., the number of parameters in the ML model. And in some real tasks, grid data can be sparse and thus a model trainable with sparse data is preferred.
    \item \textit{inference complexity - is it easy to make predictions on new inputs?} This is also related to model complexity and can be quantified by the computation complexity $O(\cdot)$ or wall-clock time.
    \item \textit{physical interpretability} - This is a novel attribute defined within this paper in Section \ref{sec:interpretability}. It focuses on the question: \textit{How does the model make predictions in a physically meaningful way?} 
    \item \textit{prediction quality - how good is the prediction (especially on new inputs)?} Metrics to quantify this include (but are not limited to) accuracy, precision, recall, F-measure for classification problems, and prediction error for regression problems. More importantly, since power grid applications are safety-critical, predictions with \textbf{low variance} are needed for regression problems, and predictions with \textbf{low false negatives (FN) and low variance} are needed for classification or detection problems. 
\end{itemize}

\subsubsection{Layer 2}

The second level of the pyramid contains two higher-level attributes: \textbf{scalability} and \textbf{generalization \& transferability.} 

\begin{itemize}
    \item \textit{scalability\cite{scalability}}: scalability focuses on \textit{how does computation complexity change as power grid size gets larger?} and whether the model is applicable on large-scale systems. This is related to basic attributes \textit{trainability} and \textit{inference complexity}.
    \item \textit{generalization\cite{generalization-DL} and transferability}: Generalization, or generalizability, is statistically quantifiable by generalization error (which is empirically often approximated by the validation/test error). Yet more broadly, it is a multi-dimensional concept to evaluate a model's performance (e.g. prediction quality) on unseen data. In the context of power grid analysis, these can be explained as follows:
    \begin{itemize}
        \item \textbf{poor generalization}: a model, once trained, is only applicable to a fixed topology or load conditions
        \item \textbf{weak generalization}: a model can work with unseen topology, generation, and load conditions.
        \item \textbf{strong generalization}: a model trained on one grid is also applicable for another. This means the operator, or a utility company, can directly buy a pre-trained model and use it.
    \end{itemize}
    In terms of transferability, due to the confidentiality of utility data and the sparse nature of grid measurements, many models in the literature are trained and validated on synthetic datasets, and can hardly guarantee the same performance when they “transfer” to the context of real-world data. The potential differences in data distributions highlight the importance of generating synthetic data that is also realistic and raises the need for evaluating the goodness of model transferability. 
\end{itemize}

\subsubsection{Layer 3} Practicality lies on the top of the pyramid. Satisfying the attributes in lower pyramid levels adds confidence that the method is practical. While this section discusses the performance evaluation of a method from ML perspective, the performance of some ML models can also be evaluated by application-level tests. For example, this paper tests the proposed warm starter by feeding its predictions into a power flow solver. A well-performing warm starter should significantly reduce the simulation time of a physics-based power flow solver.

\section{A novel warm-starter} \label{sec:method}
Motivated by the limitations of ML techniques in Section \ref{sec:ML survey}, we propose a novel physics-informed ML model to practically solve cyberattack driven contingencies on the power grid. The model is designed to be \textbf{generalizable to topology change} by using a graphical model, and \textbf{physically interpretable} by making predictions from a linear system proxy, and \textbf{scalable} by applying regularization techniques on the graphical model.

\subsection{Task definition and symbol notations} 
As Figure \ref{fig: graphical model} shows, given an input $\bm{x}$ which contains contingency information $c$ and (pre-contingency) system information $G$, a warm-starter makes prediction $\bm{y}$ which is an estimate of the post-contingency bus voltages $\bm{v}^{post}$. The model is a function mapping, which is learned from training dataset  $Data=\{(\bm{x^{(j)},\bm{y^{(j)}}})\}$, where $(j)$ denotes the $j$-th sample.


Table \ref{tab:notations} shows the symbols used in this paper.

\setlength{\tabcolsep}{6pt}
\begin{table}[htbp]
\small
\centering
	\caption{Symbols and definitions \label{tab:notations}}
	\begin{tabular}{ @{}rl@{} }  
	\toprule
	\textbf{Symbol} & \textbf{Interpretation} \\ \midrule
		$G$    
     &case data before contingency\\
     & containing topology, generation, and load settings\\
     \midrule
     $\bm{v_i}$   & the voltage at bus $i$,
     $\bm{v_i}=[v_i^{real},v_i^{imag}]^T$\\
     \midrule
     $\bm{v}^{pre/post}$ & $\bm{v}^{pre/post}=[\bm{v}^{pre/post}_1,\bm{v}^{pre/post}_2,...,\bm{v}^{pre/post}_n]^T$ \\
     &the pre/post-contingency voltages at all buses\\
     \midrule
     $c$ & contingency setting $(type, location, parameter)$\\
     &e.g. $(MadIoT, [1,3], 150\%):$ increasing loads at bus 1\\
     &and 3 to $150\%$ the original value.\\
     \midrule
     $i,n$ & bus/node index; total number of nodes\\
     $(s,t)$& a branch/edge connecting node $s$ and node $t$\\
     $\mathcal{V},\mathcal{E}$ & set of all nodes and edges: $i\in\mathcal{V},\forall i; (s,t)\in\mathcal{E}$ \\
     $j,N$ & data sample index; total number of samples\\
     \midrule
     $(\bm{x},\bm{y})$ & a sample with feature $\bm{x}$ and output $\bm{y}$\\
     &$\bm{y}=[\bm{y_1,...,y_n}]^T=[\bm{v}^{post}_1,...,\bm{v}^{post}_n]^T$ \\
	\bottomrule
	\end{tabular} 
\end{table}

\subsection{Method Overview}

\begin{figure}[h]
	\centering
	\includegraphics[width=0.6\linewidth]{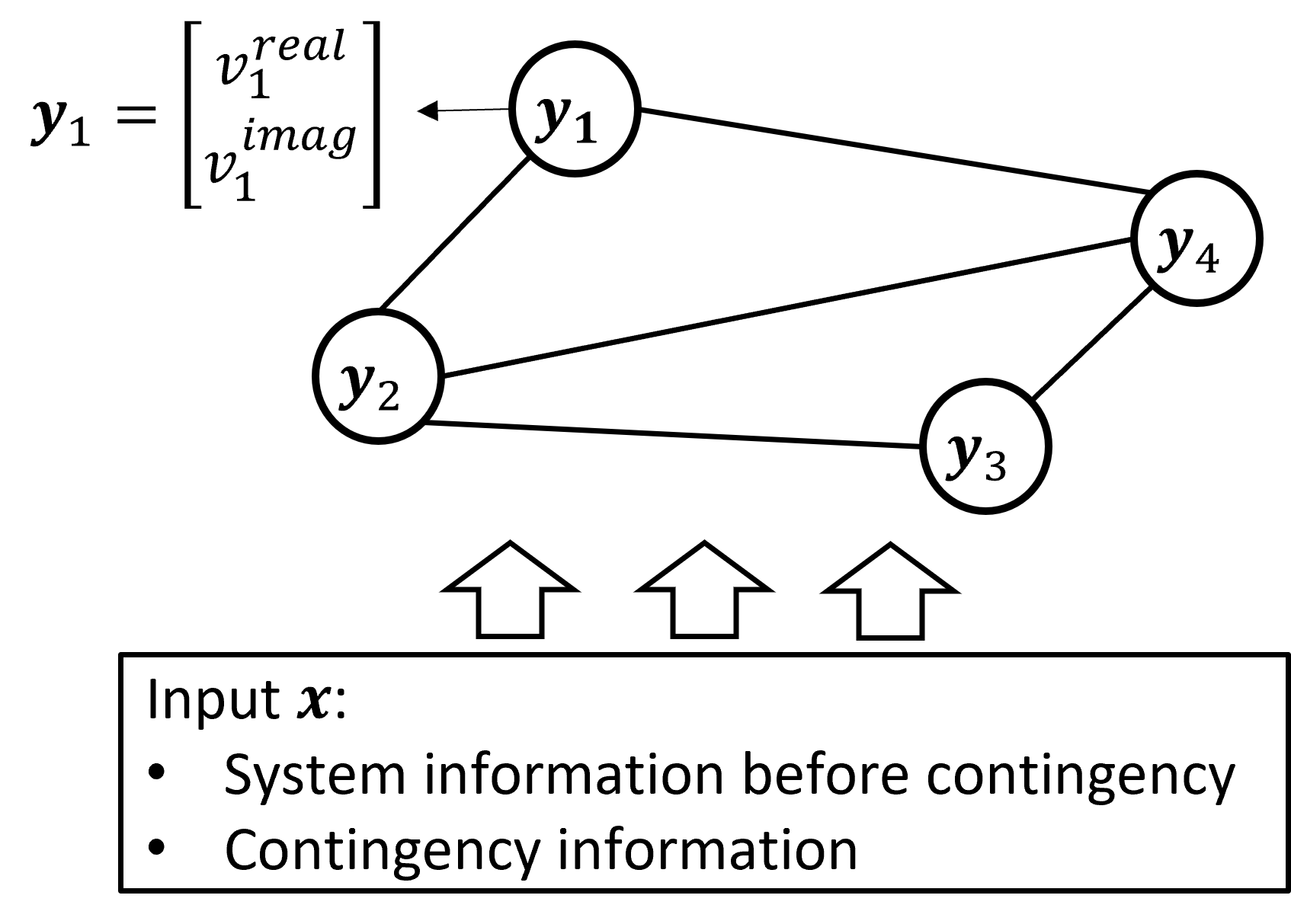}
	\caption[]{A Conditional Random Field representing a power grid during contingency: each node represents the bus voltage after contingency, each edge represents a branch status after contingency.}
	\label{fig: graphical model}
\end{figure}

Power grid can be naturally represented as a graph, as shown in Figure \ref{fig: graphical model}.
Nodes and edges on the graph correspond to physical elements on the power grid such as buses and branches (lines and transformers), respectively. Each node represents a variable $\bm{y_i}$ which denotes voltage phasor at bus $i$, whereas each edge  represents a direct inter-dependency between adjacent nodes. 

Such a graphical model enables a compact way of writing the conditional joint distribution and performing inference thereafter, using observed data. Specifically, when contingency happens, the joint distribution of variables $\bm{y}$ conditioned on input features $\bm{x}$ can be factorized in a pairwise manner\cite{MRF}, as (\ref{eq:pairwise parameterization}) shows:
\begin{align}
    &P(\bm{y}|\bm{x},\bm{\theta})=\frac{1}{Z(\bm{\theta},\bm{x})}\prod_{i=1}^n\psi_i(\bm{y_i}) \prod_{(s,t)\in\bm{E}}\psi_{st}(\bm{y_s},\bm{y_t})
    \label{eq:pairwise parameterization}
\end{align}
where $\bm{\theta}$ denotes the model parameter that maps $\bm{x}$ to $\bm{y}$; $\psi_i(\bm{y_i}),\psi_{st}(\bm{y_s},\bm{y_t})$ are node and edge potentials conditioned on $\bm{\theta}$ and $\bm{x}$; and $Z(\bm{\theta},\bm{x})$ is called the partition function that normalizes the probability values such that they sum to the value of 1.

The pairwise factorization is inspired from grid physics and provides good intuition: every edge potential records the mutual correlation between two adjacent nodes; both node and edge potentials represent their local contributions of nodes/edges to the joint distribution. 

Given a training dataset of $N$ samples $\{(\bm{x}^{(j)},\bm{y}^{(j)})\}$, the training and inference can be described briefly as:
\begin{itemize}
    \item \textbf{Training:} With proper definition of the potential functions $\psi_i(\bm{y_i}|\bm{x,\theta})$, $\psi_{st}(\bm{y_s},\bm{y_t}|\bm{x,\theta})$ (see Section \ref{sec:cGRF}-\ref{sec: NN-node and NN-edge}), the parameter $\bm{\theta}$ can be  learned by maximizing log-likelihood
   \begin{equation}
       \bm{\hat{\theta}}=arg\max_{\bm{\theta}} \sum_{j=1}^N log l(\theta)^{(j)}
   \end{equation}
   where $l(\theta)^{(j)}$ denotes the log-likelihood of the j-th sample
   \begin{equation}
       l(\theta)^{(j)}=P(\bm{y}^{(j)}|\bm{x}^{(j)},\bm{\theta})
   \end{equation}
    \item \textbf{Inference:} For any new input $\bm{x}_{test}$, we make use of the estimated parameter $\hat{\bm{\theta}}$ to make a single-point prediction 
    \begin{equation}
        \bm{\hat{y}_{test}}=arg\max_{\bm{y}} P(\bm{y|x_{test},\hat{\theta}})
        \label{eq: general inference}
    \end{equation}
\end{itemize}

The use of probabilistic graphical setting naturally integrates the domain knowledge from grid topology into the method:  

\noindent \textit{\textbf{(Domain knowledge: grid topology)} Power flow result is conditioned on the grid topology. Bus voltages of two adjacent buses connected directed by a physical linkage (line or transformer) have direct interactions.
}

Each sample in this method can have its own topology and each output is conditioned on its input topology. The following sections will further discuss how the graphical model, together with domain knowledge, enables an efficient and physically interpretable model design.

\subsection{Pairwise Conditional Gaussian Random Field} \label{sec:cGRF}
Upon representing the power grid and its contingency as a conditional pairwise MRF factorized in the form of (\ref{eq:pairwise parameterization}), we need to define the potential functions $\psi_i(\bm{y_i}),\psi_{st}(\bm{y_s},\bm{y_t})$. 

This paper builds a Gaussian random field which equivalently assumes that the output variable $(\bm{y})$ satisfies multivariate Gaussian distribution, i.e., $P(\bm{y}|\bm{x},\bm{\theta})$ is Gaussian. The justification and corresponding benefits of using Gaussian Random Field are:
\begin{itemize}
    \item partition function $Z(\bm{\theta},\bm{x})$ is easier to compute due to good statistical properties of Gaussian distribution.
    \item high physical interpretability due to a physically meaningful inference model. We will discuss this later.
\end{itemize}

Potential functions for Gaussian random field\cite{MRF} are defined as follows:
\begin{align}
    &\psi_i(\bm{y_i})=exp(-\frac{1}{2}\bm{y_i}^T\bm{\Lambda_{i}}\bm{y_i}+\bm{\eta_i}^T\bm{y_i})\label{eq:gaussian_node_potential}\\
    &\psi_{st}(\bm{y_s},\bm{y_t})=exp(-\frac{1}{2}\bm{y_s}^T\bm{\Lambda_{st}}\bm{y_t})
    \label{eq:gaussian_edge_potential}
\end{align}
By plugging \eqref{eq:gaussian_node_potential} and \eqref{eq:gaussian_edge_potential} into \eqref{eq:pairwise parameterization}, we have:
\begin{equation}
    P(\bm{y}|\bm{x},\bm{\theta})\propto exp(\bm{\eta}^T\bm{y}-\frac{1}{2}\bm{y}^T\Lambda\bm{y})
    \label{eq:gaussianMRF}
\end{equation}
where $\bm{\Lambda_{i}}$ and $\bm{\Lambda_{st}}$ parameters are the building blocks of matrix $\bm{\Lambda}$, and $\bm{\eta}$ is a column vector composed of all $\bm{\eta_i}$.
To further illustrate, consider a post-contingency grid structure in Figure \ref{fig: graphical model}. The  $\bm{\eta}$ and $\bm{\Lambda}$ matrix for this grid structure are shown in Figure \ref{fig: training process} where the $\bm{0}$ blocks in $\bm{\Lambda}$ matrix are structural zeros representing no edges at the corresponding locations.

In the model (\ref{eq:gaussianMRF}),   both $\bm{\eta}$ and $\bm{\Lambda}$ are functions of $\bm{x},\bm{\theta}$, i.e.,
\begin{equation}
    {\bm{\eta}}=f_\eta(\bm{x},\bm{{\theta_{\eta}}}), {\bm{\Lambda}}=f_\Lambda(\bm{x},\bm{{\theta_{\Lambda}}})
\end{equation}
and as expected, $P(\bm{y}|\bm{x},\bm{\theta})$ takes an equivalent form of a multivariate Gaussian distribution $N(\bm{\mu},\bm{\Sigma})$ ($\bm{\mu}$ is the mean and $\bm{\Sigma}$ is the covariance matrix) with
\begin{equation}
\bm{\eta}=\bm{\Lambda}\bm{\mu}, \bm{\Lambda}=\bm{\Sigma}^{-1}    
\end{equation}

Now based on these defined models, we seek to learn the parameter $\bm{\theta}$ through maximum likelihood estimation (MLE). The log-likelihood of each data sample can be calculated by:
\begin{equation}
    l(\bm{\theta})= logP(\bm{y}|\bm{x},\bm{\theta})= -\frac{1}{2}\bm{y}^T\bm{\Lambda}\bm{y}+\bm{\eta}^T\bm{y}-log Z(\bm{\theta},\bm{x}) \label{eq: loglikelihood}
\end{equation}
and the MLE can be written equivalently as an optimization problem that minimizes the negative log-likelihood loss on the data set of $N$ training samples:
\begin{equation}
    \min_{\bm{\theta}} -\sum_{j=1}^{N}l(\bm{\theta})^{(j)}
    \label{eq:MLE simplified}
\end{equation}

\textbf{Inference and Interpretation:}
Upon obtaining the solution of $\hat{\bm{\theta}}=[\hat{\bm{\theta_{\eta}}},\hat{\bm{\theta_{\Lambda}}}]^T$, parameters $\hat{\bm{\Lambda}}=f_\Lambda(\bm{x_{test}},\bm{\hat{\theta}})$, $\hat{\bm{\eta}}=f_\eta(\bm{x_{test}},\bm{\hat{\theta}})$ can be estimated thereafter. Then for any test contingency sample $\bm{x_{test}}$, the inference model in (\ref{eq: general inference}) is equivalent to solving $\bm{\hat{y}_{test}}$ by:
\begin{align}
    \hat{\bm{\Lambda}}\bm{\hat{y}_{test}}=\hat{\bm{\eta}}
    \label{eq: linear approx}
\end{align}

Notably, the model in (\ref{eq: linear approx}) can be seen as a \textbf{linear system approximation} of the post-contingency grid, providing a physical interpretation of the method. $\bm{\Lambda}$ is a sparse matrix with similar structure as the bus admittance matrix where the zero entries are 'structural zeros' representing that there is no branch connecting buses. $\bm{\eta}$ behaves like the net injection to the network. 

\subsection{NN-node and NN-edge}\label{sec: NN-node and NN-edge} 


Finally to implement the model, 
we need to specify the functions of  $f_\eta(\bm{x},\bm{{\theta_{\eta}}}),f_\Lambda(\bm{x},\bm{{\theta_{\Lambda}}})$. In the way we defined the potential functions, despite that the size of $\bm{\Lambda,\eta}$ increases with grid size, we can take advantage of the sparsity of $\bm{\Lambda}$ so that only some edge-wise parameters $\bm{\Lambda_{st}}$ and node-wise parameters $\bm{\Lambda_{i}},\bm{\eta_i}$ have to be learned.

\begin{figure}[h]
	\centering
	\includegraphics[width=0.95\linewidth]{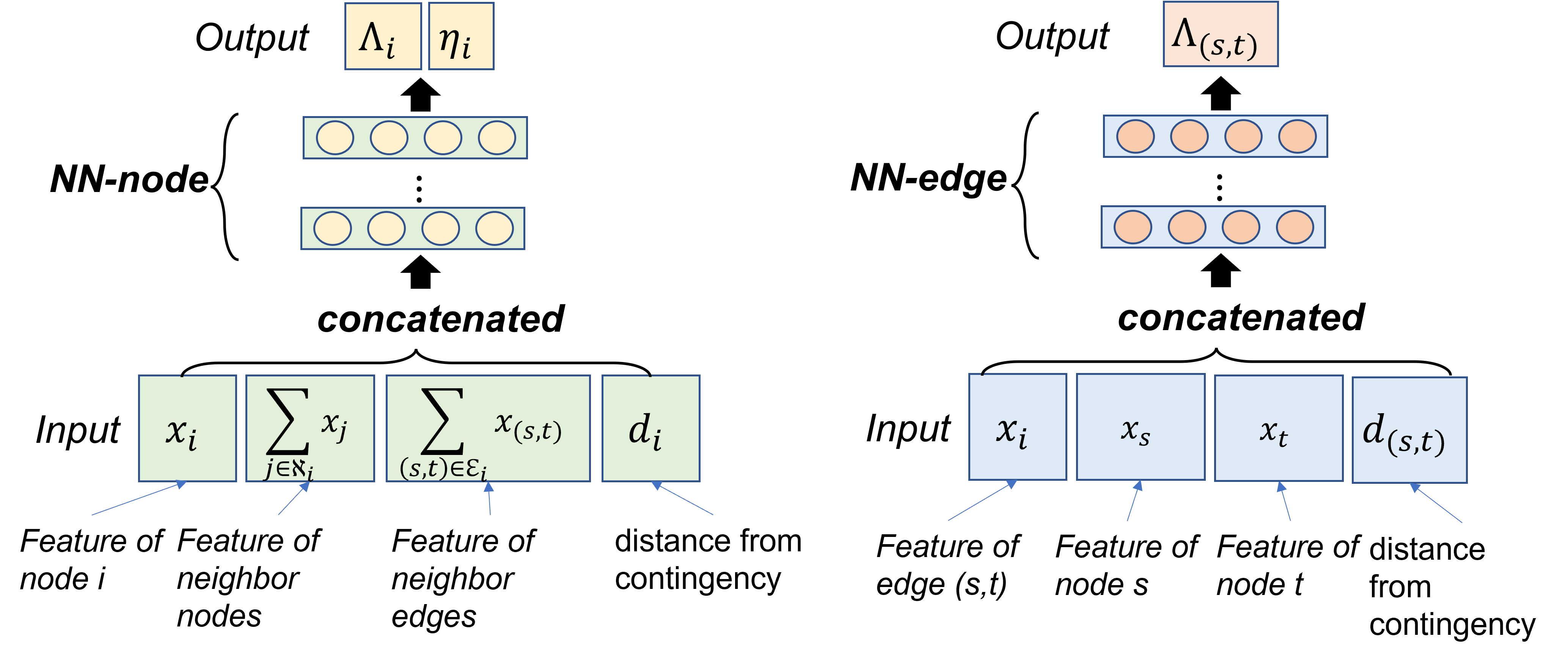}
	\caption[]{each node has a \textit{NN-node} and each edge has a \textit{NN-edge}, to map the input features to the post-contingency system characteristics.}
	\label{fig: NN-node and NN-edge}
\end{figure}

Thus the task here is to learn a function mapping from input feature to the system characteristics $\bm{\Lambda_{i}},\bm{\Lambda_{st}}, \bm{\eta_i}$ after contingency. Yet the number of $\bm{\Lambda_{i}},\bm{\Lambda_{st}}, \bm{\eta_i}$ parameters still increases with grid size, meaning that the input and output size of the model will explode for a large-scale system, requiring a much more complicated model structure and more data in training. 
To efficiently reduce the model size, this paper implements the mapping functions using local Neural Networks: each node has a \textit{NN-node} to predict $\bm{\Lambda_{i}},\bm{\eta_i}$; each edge has \textit{NN-edge} to output $\bm{\Lambda_{st}}$, as Figure \ref{fig: NN-node and NN-edge} shows.

Meanwhile, to effectively learn the mapping, \textbf{how shall we select the input features to the NN models?} To feed the most relevant features into the models, we design the input by the following domain knowledge:\\

\noindent \textit{\textbf{(Domain knowledge: decisive features)} The impact of contingency depends heavily on the importance of contingency components which can be quantified by the amount of its generation, load or power delivery.
}

\noindent \textit{\textbf{(Domain knowledge: Taylor expansion on system physics)} Let $v=h(G)$ denote any power flow simulation that maps from the case information to the voltage profile solution. By Taylor Expansion, the post-contingency voltage can be expressed as a function depending on pre-contingency system $G_{pre}$ and the system change $\Delta G$ caused by contingency:
$$\bm{v^{post}} = h(G_{pre}) + h' (G_{pre})\Delta G + \frac{1}{2}h'' (G_{pre})\Delta G^2 + ...$$
}

Therefore, the important features of pre-contingency system and system change are selected as node features to feed into the NN mappings, i.e.,
\begin{itemize}
    \item node feature $\bm{x}_i$: $v_i^{real}, v_i^{imag}, P_i, Q_i, I_i^{real}, I_i^{imag}, Q_{shunt,i}$ before contingency, and $\Delta P_{gen,i},\Delta P_{load,i},\Delta Q_{load,i}$ caused by contingency
    \item edge feature $\bm{x}_{s,t}$: admittance and shunt capacity of the 'pi-model', $\bm{x}_{s,t}=[G,B,B_{sh}]$
\end{itemize}
\subsection{Training the model with a surrogate loss}\label{sec:training}
\begin{figure}[h]
	\centering
	\includegraphics[width=0.8\linewidth]{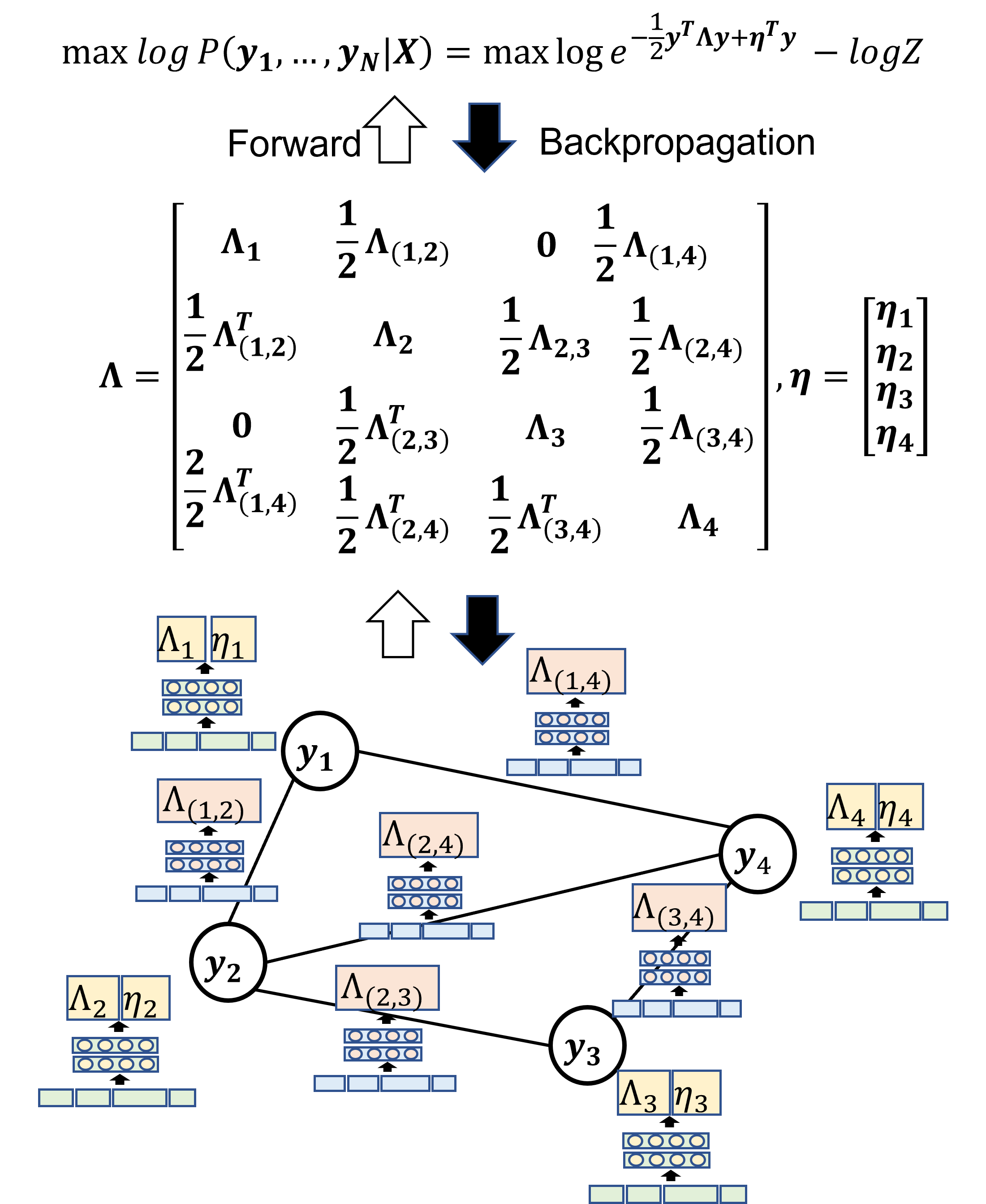}
	\caption[]{Training of the proposed method: forward pass and back-propagation.}
	\label{fig: training process}
\end{figure}

With the conditional GRF model defined in Section \ref{sec:cGRF} and the NN models designed in Section \ref{sec: NN-node and NN-edge}, the training process can be described as Figure \ref{fig: training process}, where the forward pass of NN-node and NN-edge gives $\bm{\Lambda,\eta}$, and then the loss defined from the cGRF can be calculated to further enable a backward pass that updates the parameter $\bm{\theta}$.

As written in Equation (\ref{eq: loglikelihood})-(\ref{eq:MLE simplified}), the loss function is the negative log-likelihood loss over the training data. Making use of the nice properties of Gaussian distribution, the partition function $Z(\bm{x,\theta})$ in the loss can be calculated analytically:
\begin{align}
    `   Z &= \int_{\bm{y}} exp(\bm{\eta}^T\bm{y}-\frac{1}{2}\bm{y}^T\Lambda\bm{y}) d\bm{y}\notag\\
&=\sqrt{\frac{2\pi}{|\Lambda|}} exp({\frac{\mu^T\Lambda\mu}{2}})
=\sqrt{\frac{2\pi}{|\Lambda|}} exp({\frac{\eta^T\Lambda^{-1}\eta}{2}})
    \label{eq: Z}
\end{align}

The detailed derivation can be found in Appendix \ref{appendix: Z}.

Furthermore, to enable a valid distribution $P(\bm{y}|\bm{x},\bm{\theta})$ and unique solution during inference, it is required that the $\bm{\Lambda}$ matrix is positive definite (PD), i.e., $\bm{\Lambda}\succ \bm{0}$. Therefore, adding this constraint and substituting (\ref{eq: Z}) into the loss function, the optimization problem of the proposed method can be written as:
\begin{align}
    &\min_{\bm{\theta}}\sum_{j=1}^{N}
    \frac{1}{2}\bm{y^{(j)T}}\bm{\Lambda^{(j)}}\bm{y^{(j)}}\notag\\
    &-\bm{\eta^{(j)T}}\bm{y^{(j)}}
    -\frac{1}{2}log|\bm{\Lambda^{(j)}}| 
    + \frac{1}{2}\bm{\eta^{(j)T}\Lambda^{-1(j)}\eta^{(j)}}\\
    s.t. &\notag\\
    &\text{(forward pass) }\bm{\Lambda^{(j)}}=f_\Lambda(\bm{x^{(j)}},\bm{{\theta_{\Lambda}}}),\forall j\\
    &\text{(forward pass) } \bm{\eta^{(j)}}= f_\eta(\bm{x^{(j)}},\bm{{\theta_{\eta}}}), \forall j\\
    &\text{(positive definiteness) }\bm{\Lambda^{(j)}}\succ \bm{0}, \forall j
    \label{eq: actual optimization problem}
\end{align}

In this problem, maintaining the positive definiteness of matrix $\bm{\Lambda}$ for every sample is required, not only in the final solution, but throughout the training process (due to the $log|\bm{\Lambda}|$ in the loss). Yet this can be very hard especially when the matrix size is large, making the optimization problem hard-to-solve. 

To address this issue, this paper designed a \textbf{surrogate loss} which acts as a proxy for the actual loss we wanted to minimize, and convert the problem to the following form:
\begin{align}
    &\min_{\bm{\theta}}\sum_{j=1}^{N}\frac{1}{2}\bm{(y^{(j)}-\mu^{(j)})}^T(\bm{y^{(j)}-\mu^{(j)}})\\
    s.t. &\notag\\
    &\text{(froward pass) }\bm{\Lambda^{(j)}}=f_\Lambda(\bm{x^{(j)}},\bm{{\theta_{\Lambda}}}),\forall j\\
    &\text{(forward pass) } \bm{\eta^{(j)}}= f_\eta(\bm{x^{(j)}},\bm{{\theta_{\eta}}}), \forall j\\
    &\text{(inference) } \bm{\mu^{(j)}=\Lambda^{-1(j)}\eta^{(j)}}, \forall j
    \label{eq: surrogate optimization}
\end{align}

Appendix \ref{appendix: surrogate loss} shows how the new loss mathematically approximates the original objective function. In this way, we removed the need to maintain positive-definiteness of the $\bm{\Lambda}$ in the learning process, whereas the prediction is made using a $\bm{\Lambda}$ computed from the forward pass, and thus still considers the power grid structure enforced by the graphical model. 

From decision theory, both the original and the surrogate loss aim to return an optimized model whose prediction $\bm{\hat{y}}$ approximates the ground truth $\bm{y}$, and both make predictions by finding out the linear approximation of the post-contingency system $\hat{\bm{\Lambda}}\bm{\hat{y}}=\hat{\bm{\eta}}$. 

Additionally, this surrogate optimization model can be considered as minimizing the mean squared error(MSE) loss $\frac{1}{2}||\bm{y-\hat{y}}||^2$ over the training data, where $\bm{\hat{y}}$ is the prediction (inference) made after a forward pass. 

\section{Incorporating More Physics}
\subsection{Parameter sharing: a powerful regularizer}
With each node having its own \textit{NN-node} and each edge having its \textit{NN-edge}, the number of parameters grows approximately linearly with grid size (more specifically, the number of nodes and edges). Can we reduce the model size further? The answer is yes! One option is to make all nodes share the same \textit{NN-node} and all edges share the same \textit{NN-edge}, so that there are only two NNs in total.

Why does this work? Such sharing of \textit{NN-node} and \textit{NN-edge} is an extensive use of \textbf{parameter sharing} to incorporate a domain knowledge into the network. Specially, from the physical perspective:

\noindent \textit{\textbf{(Domain knowledge: location-invariant (LI) properties)} the power grid and the impact of its contingencies have properties that are invariant to change of locations: 1) any location far enough from the contingency location will experience little local change. 2) change in any location will be governed by the same mechanism, i.e., the system equations.}

The use of parameter sharing across the grid significantly lowered the number of unique model parameters needed, and also reduced the need for a dramatic increase in training data to adequately learn the system mapping when the grid size gets larger.

\subsection{Zero-injection bus}
\noindent \textit{\textbf{(Domain knowledge: zero-injection (ZI) buses)} A bus with no generation or load connected is called a zero-injection (ZI) bus. These buses neither consume or produce power, and thus injections at these buses are zero.
}

With our model parameter $\eta$ serving as an equivalent to bus injections, we can integrate this domain knowledge by setting $\bm{\eta}_i=0$ at any ZI node $i$.





\section{Experiments}\label{sec:experiments}
\label{sec:Results}
\begin{figure*}[h]
	\centering
	\includegraphics[width=0.9\linewidth]{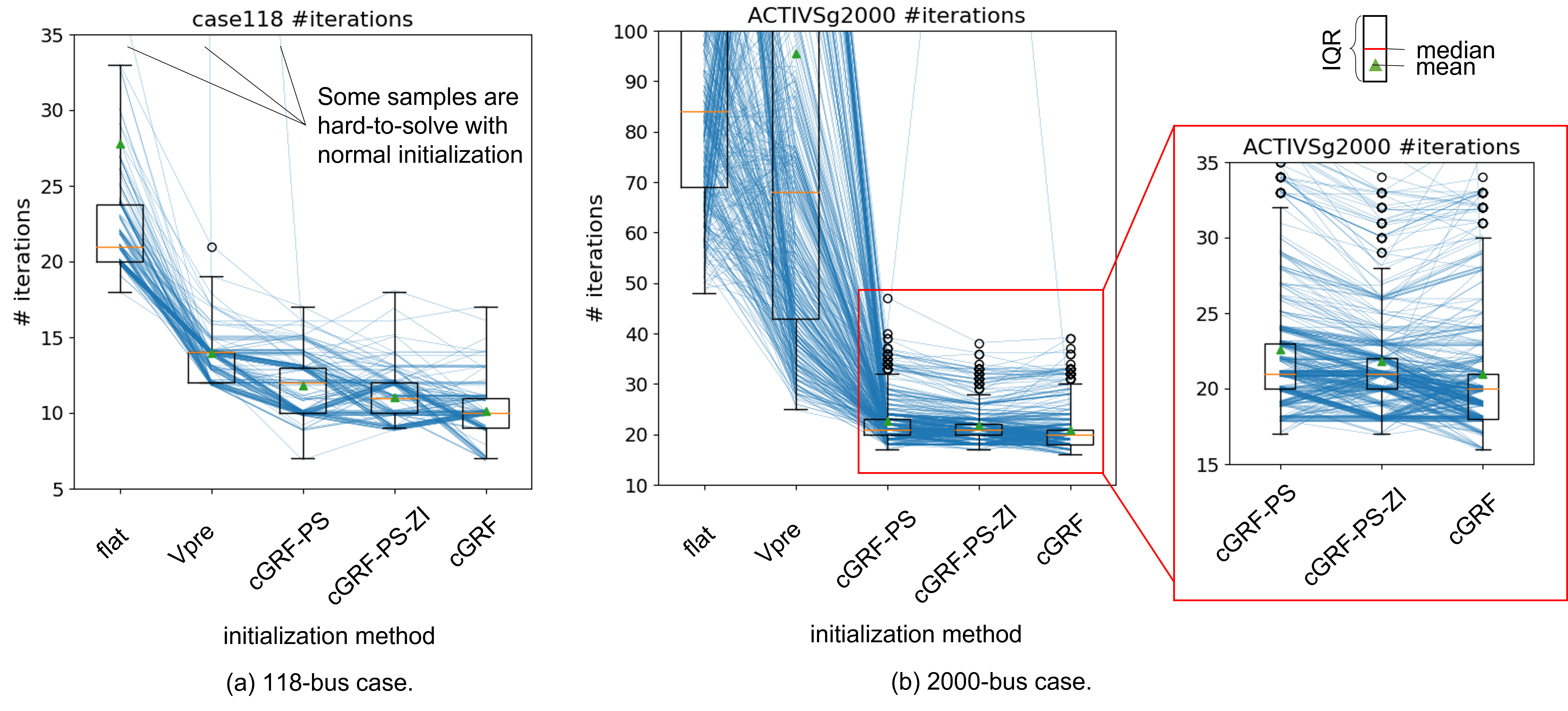}
	\caption[]{Result on test data: power flow simulation takes fewer iterations to converge with the proposed method, than traditional initialization methods  1) flat start: starting from all ones; 2) Vpre start: starting from pre-contingency voltages.}
	\label{fig:iterations case118}
\end{figure*}
\textbf{Reproducibility:} our code and data are publicly available at \url{https://github.com/ohCindy/GridWarm.git}.

This section runs experiments for a task of contingency analysis in the context of a MadIoT attack. The goal is to show the efficacy of the proposed method in predicting the post-contingency bus voltages which are further used to initialize the power flow simulation for faster convergence.

To run the experiments we use three versions of the proposed method, see Table \ref{tab:summarize methods}. These versions differ in the level of domain-knowledge that they incorporate within their model. Table \ref{tab:summarize knowledge} summarizes the domain-knowledge considered in this paper. 

\setlength{\tabcolsep}{6pt}
\begin{table}[htbp]
\small
\centering
	\caption{Summary of domain knowledge \label{tab:summarize knowledge}}
	\begin{tabular}{ @{}rll@{} }  
	\toprule
	\textbf{Knowledge} & \textbf{Technique} &\textbf{Benefits} \\ \midrule
		\textbf{topology}  
     & graphical model & - physical interpretability\\
     &  &- generalization (to topology)\\
     \midrule
     \textbf{decisive}
     & feature selection  & - accuracy \\
      \textbf{features}
     & & - physical interpretability\\
     & & - generalization (to load\&gen)\\
     \midrule
     \textbf{taylor}
     &feature selection & - accuracy \\
     \textbf{expansion}
     & & - physical interpretability\\
     \midrule
     \textbf{LI properties}
     &parameter & - trainability, scalability \\
     &sharing (PS) &- generalization ($\downarrow$ overfitting)\\
     \midrule
     \textbf{ZI bus}
     &enforce $\eta_i=0$ & - physical interpretability \\
     & & - generalization \\
	\bottomrule
	\end{tabular} 
\end{table}

\setlength{\tabcolsep}{6pt}
\begin{table}[htbp]
\small
\centering
	\caption{3 versions of the warm-starter \label{tab:summarize methods}}
	\begin{tabular}{ @{}rccc@{} }  
	\toprule
	\textbf{Knowledge \&} & \textbf{cGRF} & \textbf{cGRF-PS} & \textbf{cGRF-PS-ZI}\\ 
	\textbf{techniques} &  & & \\
	\midrule
	graphical model (cGRF) & \checkmark& \checkmark& \checkmark\\	
	\midrule
	feature selection & \checkmark&  \checkmark& \checkmark\\
	\midrule
	parameter sharing (PS) & & \checkmark& \checkmark\\
	\midrule
	ZI buses & & &\checkmark\\
	\bottomrule
	\end{tabular} 
\end{table}

\subsection{Data generation and experiment settings}

In the experiments, we evaluate the performance of our warm-starter for MadIoT-induced contingency analysis on two networks: i) IEEE 118 bus network ii) ACTIVSg2000 network. Algorithm \ref{alg: main generation} documents the three-step process used to generate synthetic data. The created data are split into train, validation, and test set by $8:1:1$.

The experiment settings that are used for the data generation, model design, and model training are documented in Table \ref{tab:experiment settings}. In our experiment, the model is designed with a very shallow 3-layer NN architecture, to save computation time, reduce overfitting, and leave room for testing whether or not a simple model design can give good performance.
\begin{algorithm}
	\caption{3-Step Data Generation Process} 
	\label{alg: main generation}
	\KwIn{Base case $G_{base}$, type of contingency $t_c$, number of data samples $N_{data}$}
	\KwOut{Generated dataset $\{(x^{(j)},y^{(j))}\}_{N_{data}}$}
	\For{{$j \gets 1$ to $N_{data}$} }{
	{\bf 1. Create a random feasible pre-contingency case $G_{pre}^{(j)}$:} each sample has random topology, generation and load level. 
	
	{\bf 2. Create contingency $c^{(j)}$ on $G_{pre}^{(j)}$: } which has attributes \textit{type, location, parameter}.
	
	{\bf 3. Simulate with droop control:} run power flow to obtain the post-contingency voltages $\bm{v}^{post}$
	}
\end{algorithm}

\setlength{\tabcolsep}{6pt}
\begin{table}[htbp]
\small
\centering
	\caption{Experiment settings \label{tab:experiment settings}}
	\begin{tabular}{ @{}rl@{} }  
	\toprule
	\textbf{Settings} & \textbf{} \\ \midrule
     $N_{data}$ & 
     case118: 1,000; ACTIVSg2000: 5,000  \\
     & split into train, val, test set by $8:1:1$\\
     \midrule
     \textit{NN-node} \& \textit{NN-edge}& shallow cylinder architecture\\
     &$(n\_layer,hidden\_ dim)=(3,64)$\\
     \midrule
     contingency $c$ & $type:$ MadIoT\\
     & $location:$ randomly sampled 50\% loads\\
     & $parameter:$ case118 200\%,\\ &  ACTIVSg2000 120\%\\
     \midrule
     optimizer & $Adam,lr=0.001, scheduler=stepLR$\\
	\bottomrule
	\end{tabular} 
\end{table}
\subsection{Physical Interpretability}
\begin{figure}[h]
	\centering
	\includegraphics[width=0.75\linewidth]{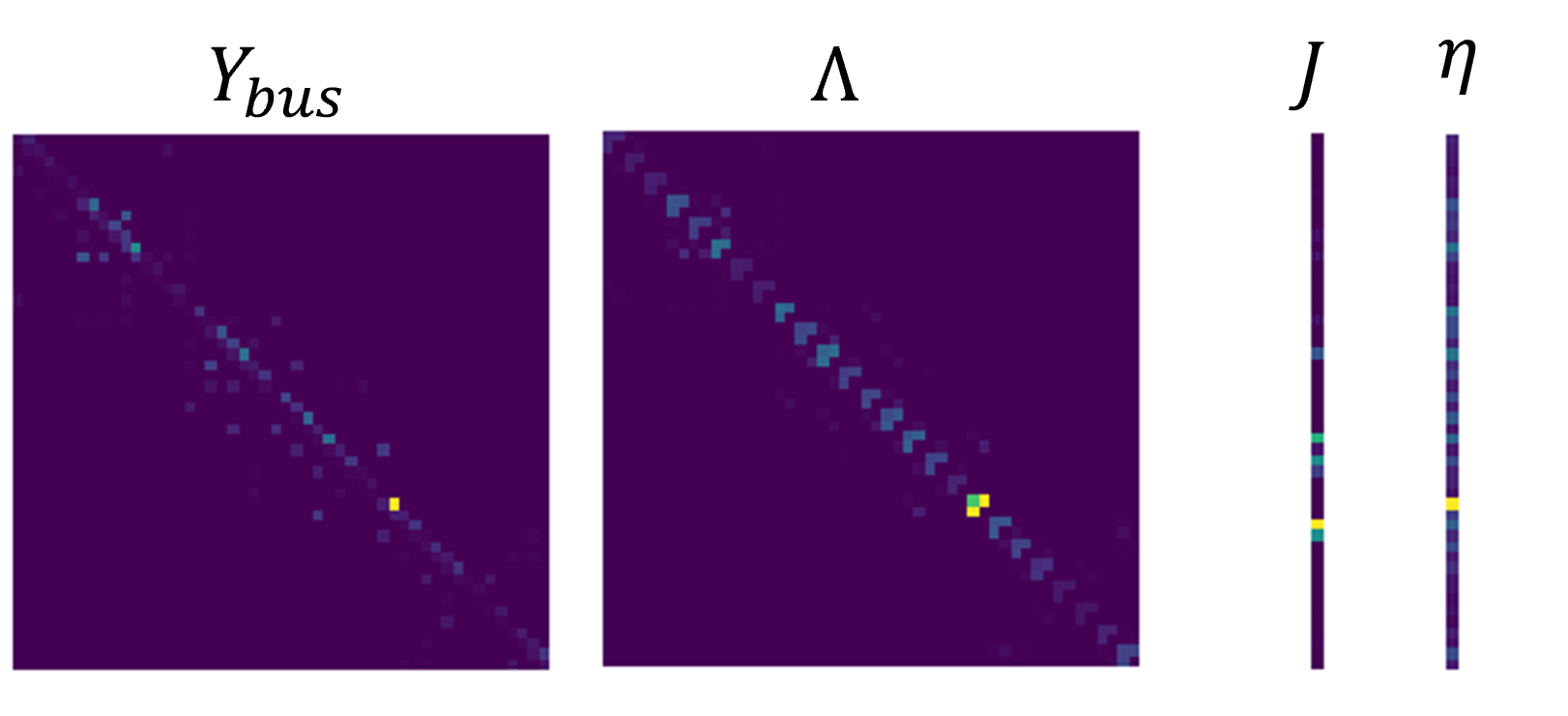}
	\caption[]{Physical interpretation: model parameters $\Lambda, \eta$ mimic the true system admittance matrix $Y_{bus}$ and injection current vector $J$.}
	\label{fig: interpretation}
\end{figure}

Figure \ref{fig: interpretation} provides insights into the physical interpretation (as discussed in Section \ref{sec:cGRF}) of our method as a linear system proxy, by comparing the model parameters $\Lambda, \eta$ and the true post-contingency system admittance matrix $Y_{bus}$ and injection current vector $J$.

\subsection{Application-level practicality}

To verify the effectiveness of the warm-starter, we compare the performance of i) power flow simulation with traditional initialization from flat start and pre-contingency solution start, and ii) power flow simulation initialized by the proposed method. 

Figure \ref{fig:iterations case118} shows the evaluation results on test data. 
We feed the ML predictions into a power flow simulator \cite{sugar-pf}. Result shows that simulation takes fewer iterations to converge with our ML initialization, when compared with traditional initialization methods. In particular, on ACTIVSg2000, many contingency samples are hard-to-solve with traditional initialization, whereas our method significantly speeds up convergence (reduces iteration by up to 70\%) even with the shallow 3-layer NN architecture.

Moreover, the \textit{lightweight} model cGRF-PS significantly reduces the total number of model parameters but achieves comparable results to the base model cGRF. And cGRF-PS-ZI further shows that a more physically meaningful solution using zero injection (ZI) knowledge can further improve convergence on the lightweight model.

\section{Conclusion}
\label{sec:Conclusion}
The main contributions of this paper include
\begin{itemize}
    \item a new concept of \textit{physical interpretability} that assesses ML models' physical meaningfulness
    \item a evaluation methodology that summarizes important attributes that can make a ML method practical for grid-specific tasks
    \item a novel warm starter for contingency analysis, with
    \begin{itemize}
        \item \textbf{generalizability to topology changes} by using a graphical model to naturally represent grid structure
        \item \textbf{physically interpretability} by generating 'global' predictions from a system-level linear proxy
        \item  \textbf{scalability} by using the graphical framework and parameter sharing techniques
    \end{itemize}
\end{itemize}

\section*{Acknowledgment}
Work in this paper is supported in part by C3.ai Inc. and Microsoft Corporation.
\bibliographystyle{IEEEtran}
\bibliography{refbib}

\appendix

\section{Appendix}


\subsection{Calculate partition function  }\label{appendix: Z}
For a multivariate Gaussian distribution $\bm{y}\sim N(\bm{\mu},\bm{\Sigma})$ where $\bm{\mu}$ denotes the mean and $\bm{\Sigma}$ denotes the covariance matrix, let $\bm{\Lambda}=\bm{\Sigma}^{-1}$, we have:
\begin{equation}
   \int_{\bm{y}} \sqrt{\frac{|\bm{\Lambda}|}{2\pi}}exp({-\frac{1}{2}(\bm{y-\mu})^T\bm{\Lambda}(\bm{y-\mu})}) d\bm{y}= 1
   \label{eq: standard Gaussian pdf integral}
\end{equation}

As mentioned earlier, the Gaussian CRF model $P(\bm{y}|\bm{x},\bm{\theta})=\frac{1}{Z(\bm{x,\theta})} exp(\bm{\eta}^T\bm{y}-\frac{1}{2}\bm{y}^T\Lambda\bm{y})$
is equivalent to a multivariate Gaussian distribution $N(\bm{\mu},\bm{\Sigma})$ with
$\bm{\eta}=\bm{\Lambda}\bm{\mu}, \bm{\Lambda}=\bm{\Sigma}^{-1}$. Thus (\ref{eq: standard Gaussian pdf integral}) can be rewritten as:
\begin{equation}
    \sqrt{\frac{|\bm{\Lambda}|}{2\pi}}exp({-\frac{\mu^T\Lambda\mu}{2}})\int_{\bm{y}} exp(\bm{\eta}^T\bm{y}-\frac{1}{2}\bm{y}^T\Lambda\bm{y}) d\bm{y}= 1
\end{equation}

Taking the nice properties of Gaussian distribution, the partition function $Z(\bm{x,\theta})$ can be calculated as:
\begin{equation}
`   Z(\bm{x,\theta}) = \int_{\bm{y}} exp(\bm{\eta}^T\bm{y}-\frac{1}{2}\bm{y}^T\Lambda\bm{y}) d\bm{y}
=\sqrt{\frac{2\pi}{|\Lambda|}} exp({\frac{\mu^T\Lambda\mu}{2}})
\end{equation}

\subsection{Surrogate loss }\label{appendix: surrogate loss}

Mathematically, due to the Gaussian distribution properties, the original optimization problem in (\ref{eq: actual optimization problem}) is equivalently:
\begin{align}
    &\min_{\bm{\theta}}\sum_{j=1}^{N}\frac{1}{2}\bm{(y^{(j)}-\mu^{(j)})}^T\bm{\Lambda^{(j)}}(\bm{y^{(j)}-\mu^{(j)}})
    -\frac{1}{2}log|\bm{\Lambda^{(j)}}| \\
    s.t. &\notag\\
    &\text{(forward pass) }\bm{\Lambda^{(j)}}=f_\Lambda(\bm{x^{(j)}},\bm{{\theta_{\Lambda}}}),\forall j\\
    &\text{(forward pass) } \bm{\eta^{(j)}}= f_\eta(\bm{x^{(j)}},\bm{{\theta_{\eta}}}), \forall j\\
    &\text{(positive definiteness) }\bm{\Lambda^{(j)}}\succ \bm{0}, \forall j\\
     &\text{(inference) } \bm{\mu^{(j)}=\Lambda^{-1(j)}\eta^{(j)}}, \forall j
\end{align}

To design a surrogate loss, we make an approximation $\bm{\Lambda^{(j)}=I}$ only in the objective function, so that $log|\bm{\Lambda}|=0$ becomes negligible.

\end{document}